\documentclass{article}

\usepackage[final]{corl_2019} 
\usepackage{graphicx}
\usepackage{times}
\usepackage{epsfig}
\usepackage{amsmath}
\usepackage{amssymb}
\usepackage{enumitem}
\usepackage{wrapfig}
\usepackage{multirow}
\usepackage{footnote}

\title{End-to-End Multi-View Fusion for 3D Object Detection in LiDAR Point Clouds}

\author{
  Yin Zhou$^{1}$
  \And
  Pei Sun$^{1}$
  \And
  Yu Zhang$^{1}$
  \thanks{Work done while at Waymo LLC.}
  \And
  Dragomir Anguelov$^{1}$
  \And
  Jiyang Gao$^{1}$
  \And
  Tom Ouyang$^{1}$
  \And
  James Guo$^{1}$
  \And
  Jiquan Ngiam$^{2}$
  \And
  Vijay Vasudevan$^{2}$ \\
}

\begin{document}
\maketitle
\setcounter{footnote}{0}
\vspace{-2em}
\begin{center}
    $^1$Waymo LLC\\
    \texttt{\small \{yinzhou, peis, yuzhangg, dragomir, jiyanggao, ouyang, guozj\}@waymo.com}\\
    $^2$Google Brain\\
    \texttt{\small \{jngiam, vrv\}@google.com}\\    
\end{center}

\begin{abstract}
Recent work on 3D object detection advocates point cloud voxelization in birds-eye view, where objects preserve their physical dimensions and are naturally separable. When represented in this view, however, point clouds are sparse and have highly variable point density, which may cause detectors difficulties in detecting distant or small objects (pedestrians, traffic signs, etc.). On the other hand, perspective view provides dense observations, which could allow more favorable feature encoding for such cases. 
In this paper, we aim to synergize the birds-eye view and the perspective view and propose a novel end-to-end multi-view fusion (MVF) algorithm, which can effectively learn to utilize the complementary information from both. Specifically, we introduce dynamic voxelization, which has four merits compared to existing voxelization methods, i) removing the need of pre-allocating a tensor with fixed size; ii) overcoming the information loss due to stochastic point/voxel dropout; iii) yielding deterministic voxel embeddings and more stable detection outcomes; iv) establishing the bi-directional relationship between points and voxels, which potentially lays a natural foundation for cross-view feature fusion. By employing dynamic voxelization, the proposed feature fusion architecture enables each point to learn to fuse context information from different views. MVF operates on points and can be naturally extended to other approaches using LiDAR point clouds.
We evaluate our MVF model extensively on the newly released Waymo Open Dataset and on the KITTI dataset and demonstrate that it significantly improves detection accuracy over the comparable single-view PointPillars baseline. 
\end{abstract}


\section{Introduction}
\label{sec:intro}

Understanding the 3D environment from LiDAR sensors is one of the core capabilities required for autonomous driving. Most techniques employ some forms of voxelization, either via custom discretization of the 3D point cloud (e.g,. Pixor~\cite{REF:yang2018pixor}) or via learned voxel embeddings (e.g., VoxelNet~\cite{REF:VoxelNet_CVPR2018}, PointPillars~\cite{REF:pointpillars_cvpr2018}). The latter typically involves pooling information across points from the same voxel, then enriching each point with context information about its neighbors. These voxelized features are then projected to a birds-eye view (BEV) representation that is compatible with standard 2D convolutions. One benefit of operating in the BEV space is that it preserves the metric space, i.e., object sizes remain constant with respect to distance from the sensor. This allows models to leverage prior information about the size of objects during training. On the other hand, as the point cloud becomes sparser or as measurements get farther away from the sensor, the number of points available for each voxel embedding becomes more limited.

Recently, there has been a lot of progress on utilizing the perspective range-image, a more native representation of the raw LiDAR data (e.g., LaserNet~\cite{REF:lasernet_CVPR2019}). This representation has been shown to perform well at longer ranges where the point cloud becomes very sparse, and especially on small objects. By operating on the ``dense'' range-image, this representation can also be very computationally efficient. Due to the perspective nature, however, object shapes are not distance-invariant and objects may overlap heavily with each other in a cluttered scene.

Many of these approaches utilize a single representation of the LiDAR point cloud, typically either BEV or range-image. As each view has its own advantages, a natural question is how to combine multiple LiDAR representations into the same model. Several approaches have looked at combining BEV laser data with perspective RGB images, either at the ROI pooling stage (MV3D~\cite{REF:Multiview3D_2017}, AVOD~\cite{REF:AVOD2018}) or at a per-point level (MVX-Net~\cite{REF:MVXNetMV_Sindagi2019}). Distinct from the idea of combining data from two different sensors, we focus on how fusing different views of the same sensor can provide a model with richer information than a single view by itself.

In this paper, we make two major contributions. First, we propose a novel end-to-end multi-view fusion (MVF) algorithm that can leverage the complementary information between BEV and perspective views of the same LiDAR point cloud. Motivated by the strong performance of models that learn to generate per-point embeddings, we designed our fusion algorithm to operate at an early stage,  where the net still preserves the point-level representation (e.g., before the final pooling layer in VoxelNet~\cite{REF:VoxelNet_CVPR2018}). Each individual 3D point now becomes the conduit for sharing information across views, a key idea that forms the basis for multi-view fusion. Furthermore, the type of embedding can be tailored for each view. For the BEV encoding, we use vertical column voxelization (\textit{i.e.,} PointPillars~\cite{REF:pointpillars_cvpr2018}) that has been shown to provide a very strong baseline in terms of both accuracy and latency. For the perspective embedding, we use a standard 2D convolutional tower on the ``range-image-like'' feature map that can aggregate information across a large receptive field, helping to alleviate the point sparsity issue. Each point is now infused with context information about its neighbors from both BEV and perspective view. These point-level embeddings are pooled one last time to generate the final voxel-level embeddings. Since MVF enhances feature learning at the point level, our approach can be conveniently incorporated to other LiDAR-based detectors~\cite{REF:VoxelNet_CVPR2018,REF:pointpillars_cvpr2018,REF:shi2019pointrcnn}.

Our second main contribution is the concept of \textit{dynamic voxelization (DV)} that offers four main benefits over traditional (i.e., \textit{hard voxelization (HV)}  ~\cite{REF:VoxelNet_CVPR2018,REF:pointpillars_cvpr2018}):
\begin{itemize}[nosep]
  \item DV eliminates the need to sample a predefined number of points per voxel. This means that every point can be used by the model, minimizing information loss.
  \item It eliminates the need to pad voxels to a predefined size, even when they have significantly fewer points. This can greatly reduce the extra space and compute overhead from HV, especially at longer ranges where the point cloud becomes very sparse. For example, previous models like VoxelNet and PointPillars allocate 100 or more points per voxel (or per equivalent 3D volume).
  \item DV overcomes stochastic dropout of points/voxels and yields deterministic voxel embeddings, which leads to more stable detection outcomes.
  \item It serves as a natural foundation for fusing point-level context information from multiple views.
\end{itemize}
MVF and dynamic voxelization allow us to significantly improve detection accuracy on the recently released Waymo Open Dataset and on the KITTI dataset. 





\section{Related Work}
\label{sec:related_works}
\textbf{2D Object Detection.} Starting from the R-CNN \cite{REF:girshick2014rich} detector proposed by Girshick \textit{et al.}, researchers have developed many modern detector architectures based on Convolutional Neural Networks (CNN). Among them, there are two representative branches: two-stage detectors \cite{REF:FasterRCNN_NIPS2015, REF:FastRCNN_ICCV2015} and single-stage detectors \cite{REF:redmon2016you, REF:SSD_ECCV2016, REF:Redmon2017YOLO9000BF}. The seminal Faster RCNN paper~\cite{REF:FasterRCNN_NIPS2015} proposes a two-stage detector system, consisting of a Region Proposal Network (RPN) that produces candidate object proposals and a second stage network, which processes these proposals to predict object classes and regress bounding boxes. On the single-stage detector front, SSD by Liu~\textit{et al.}~\cite{REF:SSD_ECCV2016} simultaneously classifies which anchor boxes among a dense set contain objects of interest, and regresses their dimensions.  Single-stage detectors are usually more efficient than two-stage detectors in terms of inference time, but they achieve slightly lower accuracy compared to their two-stage counterparts on the public benchmarks such as MSCOCO~\cite{REF:lin2014microsoft}, especially on smaller objects. Recently Lin \textit{et al.} demonstrated that using the focal loss function \cite{REF:FocalLoss_PAMI2018} on a single-stage detector can lead to superior performance than two-stage methods, in terms of both accuracy and inference time.

\textbf{3D Object Detection in Point Clouds.} A popular paradigm for processing a point cloud produced by LiDAR is to project it in birds-eye view (BEV) and transform it into a multi-channel 2D pseudo-image, which can then be processed by a 2D CNN architecture for both 2D and 3D object detection. The transformation process is usually hand-crafted, some representative works include Vote3D~\cite{REF:VotingforVoting_RSS2015}, Vote3Deep~\cite{REF:Vote3Deep_ICRA2017}, 3DFCN~\cite{REF:3DFCN_RSJ2017}, AVOD \cite{REF:ku2018joint}, PIXOR \cite{REF:yang2018pixor} and Complex YOLO \cite{REF:simony2018complex}. VoxelNet by Zhou \textit{et al.}~ \cite{REF:VoxelNet_CVPR2018} divides the point cloud into a 3D voxel grid (\textit{i.e.} voxels) and uses a PointNet-like network \cite{REF:Qi2017PointNetDL} to learn an embedding of the points inside each voxel. PointPillars \cite{REF:pointpillars_cvpr2018} builds on the idea of VoxelNet to encode the points feature on pillars (\textit{i.e.} vertical columns). Shi~\textit{et al.} \cite{REF:shi2019pointrcnn} propose a PointRCNN model that utilizes a two-stage pipeline, in which the first stage produces 3D bounding box proposals and the second stage refines the canonical 3D boxes. Perspective view is another widely used representation for LiDAR. Along this line of research, some representative works are VeloFCN~\cite{REF:VeloFCN2016} and LaserNet~\cite{REF:lasernet_CVPR2019}.

\textbf{Multi-Modal Fusion.} Beyond using only LiDAR, MV3D~\cite{REF:Multiview3D_2017} combines CNN features extracted from multiple views (front view, birds-eye view as well as camera view) to improve 3D object detection accuracy. A separate line of work, such as Frustum PointNet~\cite{REF:qi2017frustum} and PointFusion~\cite{REF:pointfusion_CVPR2018}, first generates 2D object proposals from the RGB image using a standard image detector and extrudes each 2D detection box to a 3D frustum, which is then processed by a PointNet-like network\cite{REF:Qi2017PointNetDL,REF:Pointnet++NIPS2017} to predict the corresponding 3D bounding box. ContFuse~\cite{REF:ContFuse_ECCV2018} combines discrete BEV feature map with image information by interpolating RGB features based on 3D point neighborhood. HDNET~\cite{REF:HDNET_CoRL2018} encodes elevation map information together with BEV feature map. MMF~\cite{REF:MMF_ATG_cvpr2019} fuses BEV feature map, elevation map and RGB image via multi-task learning to improve detection accuracy. Our work introduces a method for point-wise feature fusion that operates at the point-level rather than the voxel or ROI level. This allows it to better preserve the original 3D structure of the LiDAR data, before the points have been aggregated via ROI or voxel-level pooling.


\section{Multi-View Fusion}
\label{sec:proposed_method}
Our Multi-View Fusion (MVF) algorithm consists of two novel components: dynamic voxelization and feature fusion network architecture. We introduce each in the following subsections.

\subsection{Voxelization and Feature Encoding}
\label{subsec:dynamic_voxelization}
\begin{figure}
    \centering
    \includegraphics[width=0.8\textwidth]{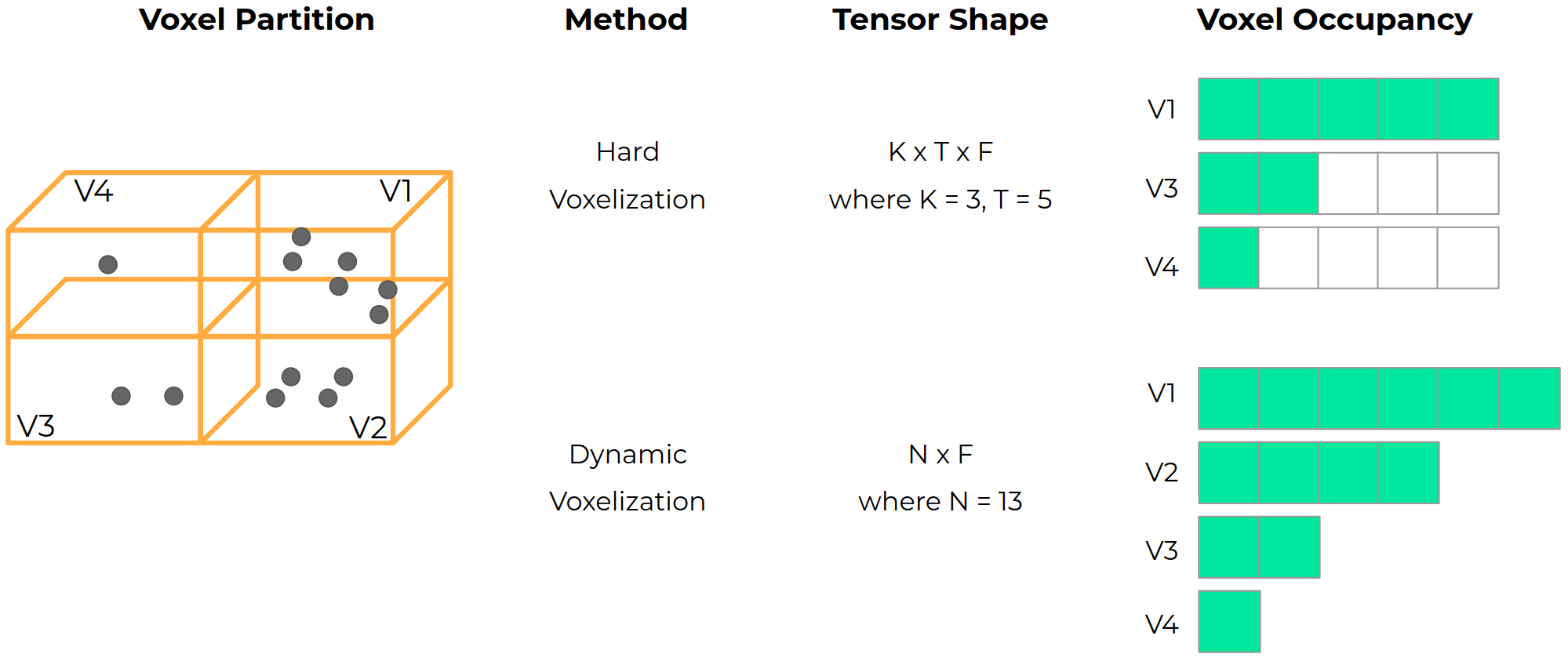}
    \caption{Illustration of the differences between \textit{hard voxelization} and \textit{dynamic voxelization}. The space is devided into four voxels, indexed as $\mathbf{v}_1$, $\mathbf{v}_2$, $\mathbf{v}_3$, $\mathbf{v}_4$, which contain 6, 4, 2 and 1 points respectively. \textit{hard voxelization} drops one point in $\mathbf{v}_1$ and misses $\mathbf{v}_2$, with $15F$ memory usage, whereas \textit{dynamic voxelization} captures all four voxels with optimal memory usage $13F$.}
    \label{fig:hard_vs_dynamic}
\end{figure}

Voxelization divides a point cloud into an evenly spaced grid of voxels, then generates a many-to-one mapping between 3D points and their respective voxels. VoxelNet~\cite{REF:VoxelNet_CVPR2018} formulates voxelization as a two stage process: \textit{grouping} and \textit{sampling}. Given a point cloud $\mathbf{P} = \{\mathbf{p}_1, \ldots,\mathbf{p}_N\}$, the process assigns $N$ points to a buffer with size $K \times T \times F$, where $K$ is the maximum number of voxels, $T$ is the maximum number of points in a voxel and $F$ represents the feature dimension. In the \textit{grouping} stage, points $\{\mathbf{p}_i\}$ are assigned to voxels $\{\mathbf{v}_j\}$ based on their spatial coordinates. Since a voxel may be assigned more points than its fixed point capacity $T$ allows, the \textit{sampling} stage sub-samples a fixed $T$ number of points from each voxel. Similarly, if the point cloud produces more voxels than the fixed voxel capacity $K$, the voxels are sub-sampled. On the other hand, when there are fewer points (voxels) than the fixed capacity $T$ ($V$), the unused entries in the buffer are zero-padded. We call this process \textit{hard voxelization}~\cite{REF:VoxelNet_CVPR2018}.

Define $F_{V}(\mathbf{p}_i)$ as the mapping that assigns each point $\mathbf{p}_i$ to a voxel $\mathbf{v}_j$ where the point resides and define $F_{P}(\mathbf{v}_j)$ as the mapping that gathers points within a voxel $\mathbf{v}_j$. Formally, \textit{hard voxelization} can be summarized as
\begin{align}
    F_{V}(\mathbf{p}_i) & = 
    \begin{cases}
        \emptyset & \text{$\mathbf{p}_i$ or $\mathbf{v}_j$ is dropped out} \\
        \mathbf{v}_j & \text{otherwise} 
    \end{cases} \\ 
    F_{P}(\mathbf{v}_j) & = 
    \begin{cases}
        \emptyset & \text{$\mathbf{v}_j$ is dropped out} \\
        \{\mathbf{p}_i\ \mid \forall \mathbf{p}_i \in \mathbf{v}_j\} & \text{otherwise} 
    \end{cases}   
\end{align}
\textit{Hard voxelization (HV)} has three intrinsic limitations: (1) As points and voxels are dropped when they exceed the buffer capacity, HV forces the model to throw away information that may be useful for detection; (2) This stochastic dropout of points and voxels may also lead to non-deterministic voxel embeddings, and consequently unstable or jittery detection outcomes; (3) Voxels that are padded cost unnecessary computation, which hinders the run-time performance.

\begin{figure}[t!] 
    \centering
    \includegraphics[width=0.9\textwidth]{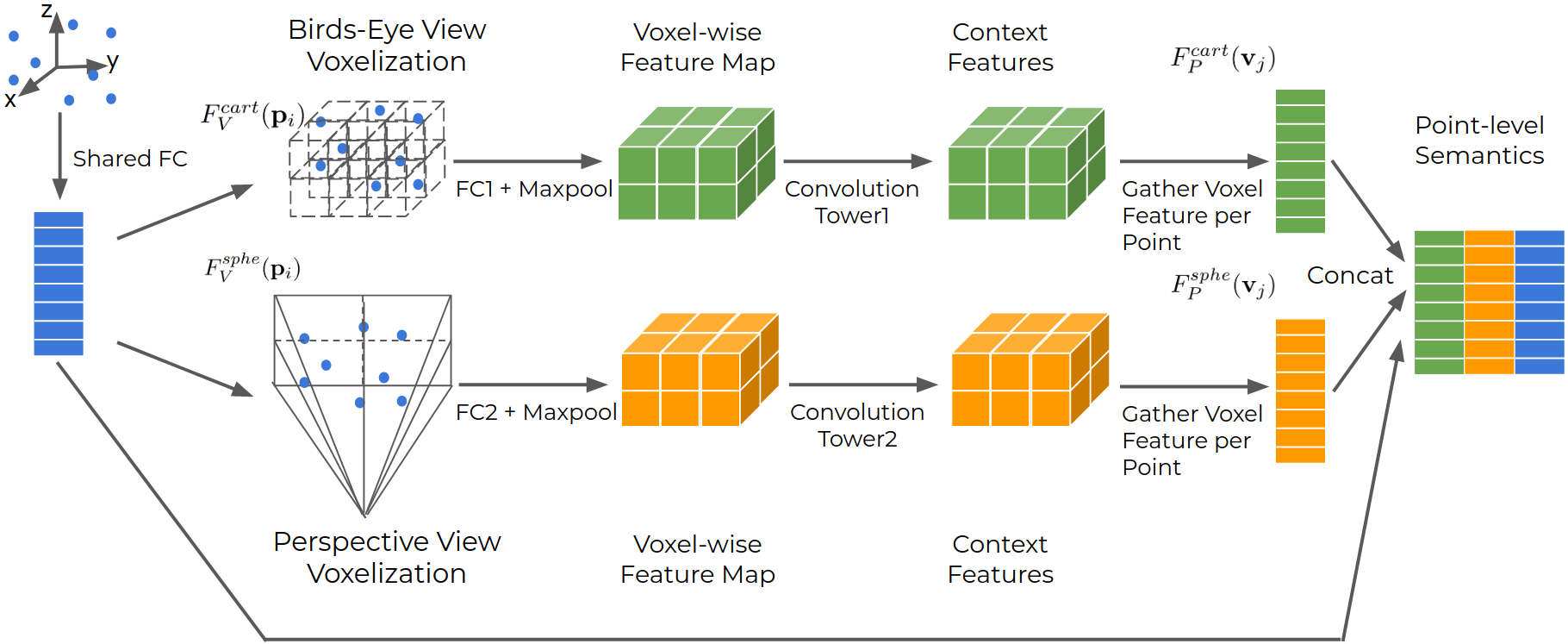}
    \caption{Multi-View Fusion (MVF) Network Architecture. Given a raw LiDAR point cloud as input, the proposed MVF first embeds each point into a high dimensional feature space via one fully connected (FC) layer, which is shared for different views. Then, it applies \textit{dynamic voxelization} in the birds-eye view and the perspective view respectively and establishes the bi-directional mapping ($F_{V}^*(\mathbf{p}_i)$ and $F_{P}^*(\mathbf{v}_j)$) between points and voxels therein, where $* \in \{cart, sphe\}$. Next, in each view, it employs one additional FC layer to learn view-dependent features, and by referencing $F_{V}^*(\mathbf{p}_i)$ it aggregates voxel information via Max Pooling. Over the voxel-wise feature map, it uses a convolution tower to further process context information within an enlarged receptive field, while still maintaining the same spatial resolution. Finally, based on $F_{P}^*(\mathbf{v}_j)$, it fuses features from three different sources for each point, \textit{i.e.,} the corresponding voxel features from the birds-eye view and the perspective view as well as the corresponding point feature obtained via the shared FC.}
    \label{fig:fusion_architecture}
\end{figure}

We introduce \textit{dynamic voxelization (DV)} to overcome these drawbacks. DV keeps the \textit{grouping} stage the same, however, instead of sampling the points into a fixed number of fixed-capacity voxels, it preserves the complete mapping between points and voxels. As a result, the number of voxels and the number of points per voxel are both dynamic, depending on the specific mapping function. This removes the need for a fixed size buffer and eliminates stochastic point and voxel dropout. The point-voxel relationships can be formalized as
\begin{align}
    F_{V}(\mathbf{p}_i) & = \mathbf{v}_j,  \forall i\\
    F_{P}(\mathbf{v}_j) & = \{\mathbf{p}_i\ \mid \forall \mathbf{p}_i \in \mathbf{v}_j\}, \forall j
\end{align}
Since all the raw point and voxel information is preserved, \textit{dynamic voxelization} does not introduce any information loss and yields deterministic voxel embeddings, leading to more stable detection results. In addition, $F_{V}(\mathbf{p}_i)$ and $F_{P}(\mathbf{v}_j)$ establish bi-directional relationships between every pair of $\mathbf{p}_i$ and $\mathbf{v}_j$, which lays a natural foundation for fusing point-level context features from different views, as will be discussed shortly.

Figure~\ref{fig:hard_vs_dynamic} illustrates the key differences between \textit{hard voxelization} and \textit{dynamic voxelization}. In this example, we set $K=3$ and $T=5$ as a balanced trade off between point/voxel coverage and memory/compute usage. This still leaves nearly half of the buffer empty. Moreover, it leads to points dropout in the voxel $\mathbf{v}_1$ and a complete miss of the voxel $\mathbf{v}_2$, as a result of the random sampling. To have full coverage of the four voxels, \textit{hard voxelization} requires at least $4 \times 6 \times F$ buffer size. Clearly, for real-world LiDAR scans with highly variable point density, achieving a good balance between point/voxel coverage and efficient memory usage will be a challenge for \textit{hard voxelization}.
On the other hand, \textit{dynamic voxelization} dynamically and efficiently allocates resources to manage all points and voxels. In our example, it ensures the full coverage of the space with the minimum memory usage of $13F$. Upon completing voxelization, the LiDAR points can be transformed into a high dimensional space via the feature encoding techniques reported in ~\cite{REF:Qi2017PointNetDL,REF:VoxelNet_CVPR2018,REF:pointpillars_cvpr2018}. 



\subsection{Feature Fusion}
\label{subsec:fusion_network}
\begin{wrapfigure}{r}{0.4\textwidth}
\vspace{-2em}
  \begin{center}
    \includegraphics[width=0.4\textwidth]{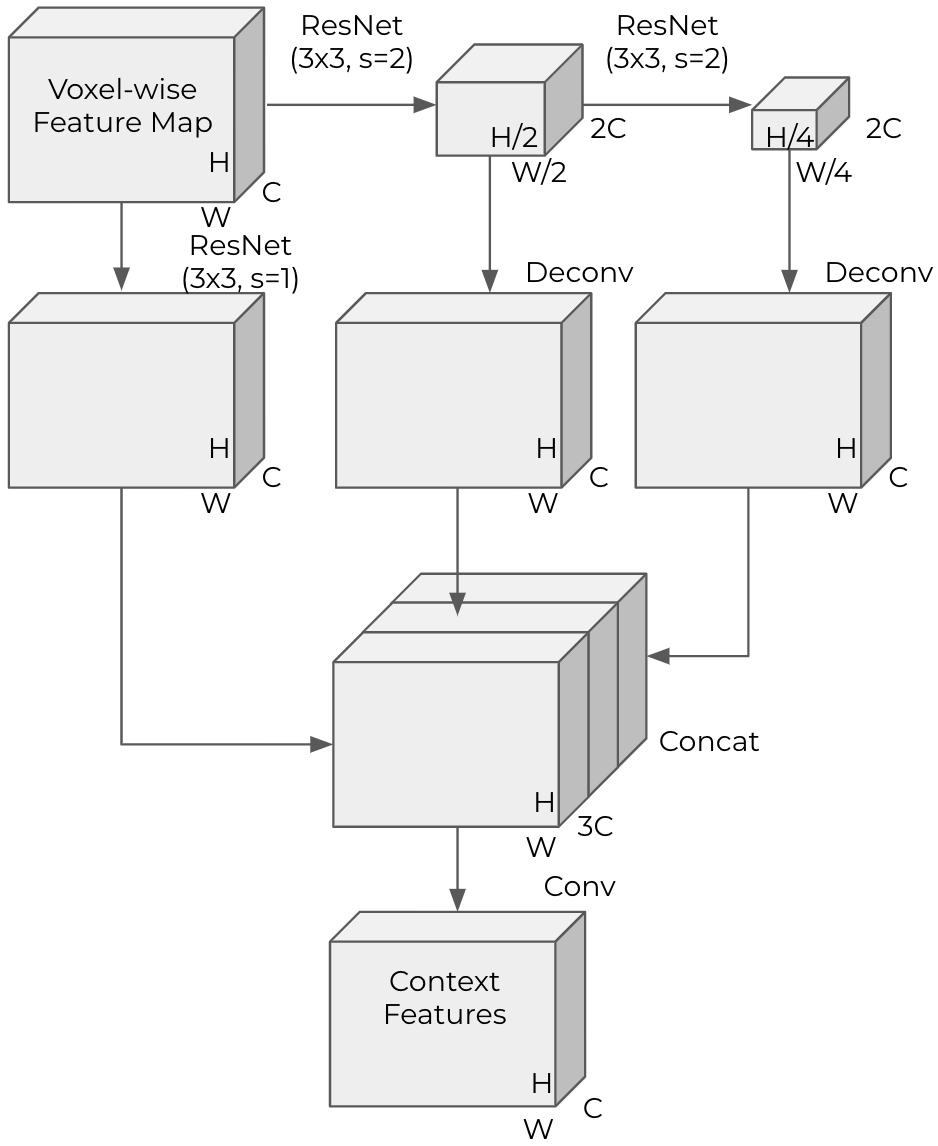}
  \end{center}
  \vspace{-1em}
  \caption{Convolution tower for encoding context information.}
  \vspace{-10pt}
  \label{fig:view_backbone}
\end{wrapfigure}
\paragraph{Multi-View Representations.} Our aim is to effectively fuse information from different views based on the same LiDAR point cloud.
We consider two views: the birds-eye view and the perspective view. The birds-eye view is defined based on the Cartesian coordinate system, in which objects preserve their canonical 3D shape information and are naturally separable. The majority of current 3D object detectors~\cite{REF:VoxelNet_CVPR2018,REF:pointpillars_cvpr2018} with \textit{hard voxelization} operate in this view. However it has the downside that the point cloud becomes highly sparse at longer ranges.  On the other hand, the perspective view can represent the LiDAR range image densely, and can have a corresponding tiling of the scene in the Spherical coordinate system. The shortcoming of perspective view is that object shapes are not distance-invariant and objects can overlap heavily with each other in a cluttered scene. Therefore, it is desirable to utilize the complementary information from both views.

So far, we have considered each voxel as a cuboid-shaped volume in the birds-eye view. Here, we propose to extend the conventional voxel to a more generic idea, in our case, to include a 3D frustum in perspective view. Given a point cloud $\{(x_i, y_i, z_i) \mid i = 1, \ldots, N\}_{cart}$ defined in the Cartesian coordinate system, its Spherical coordinate representation is computed as
\begin{equation}
    \{(\varphi_i, \theta_i, d_i) \mid \varphi_i = \arctan(\frac{y_i}{x_i}), \theta_i = \arccos(\frac{z_i}{d_i}), d_i = \sqrt{x_i^2 + y_i^2 + z_i^2}, i = 1, \ldots, N\}_{sphe}.
\end{equation}
For a LiDAR point cloud, applying \textit{dynamic voxelization} in both the birds-eye-view and the perspective view will expose each point within different local neighborhoods, \textit{i.e.,} Cartesian voxel and Spherical frustum, thus allow each point to leverage the complementary context information. The established point/voxel mappings are ($F_{V}^{cart}(\mathbf{p}_i)$, $F_{P}^{cart}(\mathbf{v}_j)$) and ($F_{V}^{sphe}(\mathbf{p}_i)$, $F_{P}^{sphe}(\mathbf{v}_j)$) for the birds-eye view and the perspective view, respectively. 

\paragraph{Network Architecture} As illustrated in Fig.~\ref{fig:fusion_architecture}, the proposed MVF model takes the raw LiDAR point cloud as input. First, we compute point embeddings. For each point, we compute its local 3D coordinates in the voxel or frustum it belongs to. The local coordinates from the two views and the point intensity are concatenated before they are embedded into a 128D feature space via one fully connected (FC) layer. The FC layer is composed of a linear layer, a batch normalization (BN) layer and a rectified linear unit (ReLU) layer. Then, we apply \textit{dynamic voxelization} in the both the birds-eye view and the perspective view and establish the bi-directional mapping ($F_{V}^*(\mathbf{p}_i)$ and $F_{P}^*(\mathbf{v}_j)$) between points and voxels, where $* \in \{cart, sphe\}$. Next, in each view, we employ one additional FC layer to learn view-dependent features with 64 dimensions, and by referencing $F_{V}^*(\mathbf{p}_i)$ we aggregate voxel-level information from the points within each voxel via max pooling. Over this voxel-level feature map, we use a convolution tower to further process context information, in which the input and output feature dimensions are both 64. Finally, using the point-to-voxel mapping $F_{P}^*(\mathbf{v}_j)$, we fuse features from three different information sources for each point: 1) the point's corresponding Cartesian voxel from the birds-eye view, 2) the point's corresponding Spherical voxel from the perspective view, and 3) the point-wise features from the shared FC layer. The point-wise feature can be optionally transformed to a lower feature dimension to reduce computational cost.

The architecture of the convolution tower is shown in Figure~\ref{fig:view_backbone}. We apply two ResNet layers~\cite{REF:Resnet2016}, each with $3 \times 3$ 2D convolution kernels and stride size $2$, to gradually downsample the input voxel feature maps into tensors with $1/2$ and $1/4$ of the original feature map dimensions. Then, we upsample and concatenate these tensors to construct a feature map with the same spatial resolution as the input. Finally, this tensor is transformed to the desired feature dimension. Note that the consistent spatial resolution between input and output feature maps effectively ensures that the point/voxel correspondences remain unchanged.
\subsection{Loss Function}
\label{subsec:loss_function}
We use the same loss functions as in SECOND~\cite{REF:second_2018} and PointPillars~\cite{REF:pointpillars_cvpr2018}. We parametrize ground truth and anchor boxes as $(x^g, y^g, z^g, l^g, w^g, h^g, \theta^g)$ and $(x^a, y^a, z^a, l^a, w^a, h^a, \theta^a)$ respectively. The regression residuals between ground truth and anchors are defined as:
\begin{align}
    & \Delta_x  = \frac{x^g - x^a}{d^a}, \Delta_y = \frac{y^g - y^a}{d^a}, \Delta_z = \frac{z^g - z^a}{h^a},\\
    & \Delta_l  = \log\frac{l^g}{l^a}, \Delta_w = \log\frac{w^g}{w^a}, \Delta_h = \log\frac{h^g}{h^a},\\
    & \Delta_\theta  = \theta^g - \theta^a
\end{align}
where $d^a = \sqrt{(l^a)^2 + (w^a)^2}$ is the diagonal of the base of the anchor box~\cite{REF:VoxelNet_CVPR2018}. The overall regression loss is:
\begin{equation}
    L_{\textrm{reg}} = \textrm{SmoothL1} (\sin(\tilde{\Delta_\theta} - \Delta_\theta)) + 
    \sum\limits_{r \in \{\Delta_x, \Delta_y, \Delta_z, \Delta_l, \Delta_w, \Delta_h\}} \textrm{SmoothL1} (\tilde{r} - r)
\end{equation}
where $\tilde{*}$ denotes predicted residuals. For anchor classification, we use the focal loss~\cite{REF:FocalLoss_PAMI2018}:
\begin{equation}
    L_{\textrm{cls}} = -\alpha (1 - p)^\gamma \log p
\end{equation}
where p denotes the probability as a positive anchor. We adopt the recommended configurations from \cite{REF:FocalLoss_PAMI2018} and set $\alpha = 0.25$ and $\gamma = 2$.

During training, we use the Adam optimizer~\cite{REF:Adam} and apply cosine decay to the learning rate. The initial learning rate is set to $1.33 \times 10^{-3}$ and ramps up to $1.5 \times 10^{-3}$ during the first epoch. The training finishes after 100 epochs.

	
\section{Experimental Results}
\label{sec:experiments}
To investigate the effectiveness of the proposed MVF algorithm, we have reproduced a recently published top-performing algorithm, PointPillars~\cite{REF:pointpillars_cvpr2018}, as our baseline. PointPillars is a LiDAR-based single-view 3D detector using \textit{hard voxelization}, which we denote as \textit{HV+SV} in the results. In fact, PointPillars can be conveniently summarized as three functional modules: voxelization in the birds-eye view, point feature encoding and a CNN backbone. To more directly examine the importance of \textit{dynamic voxelization}, we implement a variant of PointPillars by using dynamic instead of hard voxelization, which we denote \textit{DV+SV}. Finally, our MVF method features both the proposed \textit{dynamic voxelization} and multi-view feature fusion network.
For a fair comparison, we keep the original PointPillars network backbone for all three algorithms: we learn a 64D point feature embedding for \textit{HV+SV} and \textit{DV+SV} and reduce the output dimension of MVF to 64D, as well. 

\subsection{Evaluation on the Waymo Open Dataset}
\label{sec:od}
\textbf{Dataset.}
We have tested our method on the Waymo Open Dataset, which is a large-scale dataset recently released for benchmarking object detection algorithms at industrial production level. 
\begin{figure}[t!]
    \centering
    \includegraphics[width=1.0\textwidth]{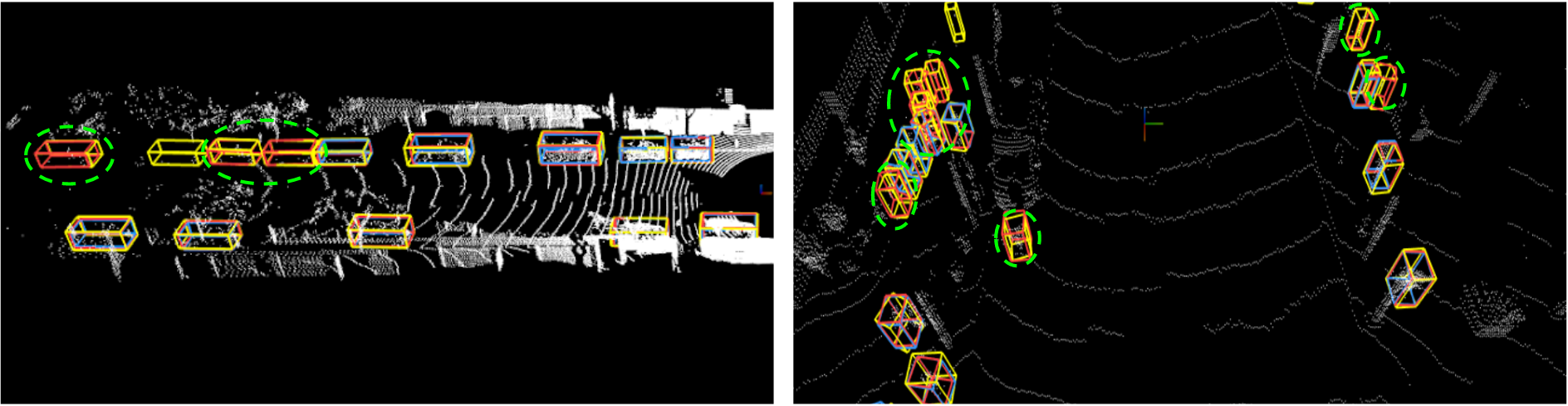}
    \caption{Visual comparison between \textit{DV+SV} and MVF on the Waymo Open Dataset. Color schemes are: grouth truth: yellow, \textit{DV+SV}: blue, MVF: red. Missing detections by \textit{DV+SV} are highlighed in green dashed circles. Best viewed in color.}
    \label{fig:error_analysis}
\end{figure}
The dataset provides information collected from a set of sensors on an autonomous vehicle, including multiple LiDARs and cameras. It captures multiple major cities in the U.S., under a variety of weather conditions and across different times of the day. 
The dataset provides a total number of 1000 sequences. Specifically, the training split consists of 798 sequences of 20s duration each, sampled at 10Hz, containing 4.81M vehicle and 2.22M pedestrian boxes. The validation split consists of 202 sequences with the same duration and sampling frequency, containing 1.25M vehicle and 539K pedestrian boxes. 
The effective annotation radius is 75m for all object classes. For our experiments, we evaluate both 3D and BEV object detection metrics for vehicles and pedestrians. 

Compared to the widely used KITTI dataset~\cite{REF:kitti_CVPR2012}, the Waymo Open Dataset has several advantages: (1) It is more than 20 times larger than KITTI, which enables performance evaluation at a scale that is much closer to production; (2) It supports detection for the full 360-degree field of view (FOV), unlike the 90-degree forward FOV for KITTI. (3) Its evaluation protocol considers realistic autonomous driving scenarios including annotations within the full range and under all occlusion conditions, which makes the benchmark substantially more challenging.

\begin{table}[h!]
\begin{center}
\begin{tabular}{|l|c|c|c|c|c|c|c|c|}
\hline
\multirow{2}{*}{Method} & \multicolumn{4}{c|}{BEV AP (IoU=0.7)}              & \multicolumn{4}{c|}{3D AP (IoU=0.7)}               \\ \cline{2-9}
            & Overall & 0 - 30m  & 30 - 50m & 50m - Inf & Overall & 0 - 30m & 30 - 50m & 50m - Inf \\ \hline
HV+SV & 75.57 & 92.1 & 74.06 & 55.47 & 56.62 & 81.01 & 51.75 & 27.94 \\ \hline

DV+SV & 77.18 & 93.04 & 76.07 & 57.67 & 59.29 & 84.9 & 56.08 & 31.07 \\ \hline

MVF & \textbf{80.40} & \textbf{93.59} & \textbf{79.21} & \textbf{63.09} & \textbf{62.93} & \textbf{86.30} & \textbf{60.02} & \textbf{36.02} \\ \hline
\end{tabular}
\\
\end{center}
\caption{Comparison of methods for vehicle detection on the Waymo Open Dataset.}
\label{eval-vehicle}
\vspace{-2em}
\end{table}

\begin{table}[h!]
\begin{center}
\begin{tabular}{|l|c|c|c|c|c|c|c|c|}
\hline
\multirow{2}{*}{Method} & \multicolumn{4}{c|}{BEV AP (IoU=0.5)}              & \multicolumn{4}{c|}{3D AP (IoU=0.5)}               \\ \cline{2-9}
            & Overall & 0 - 30m & 30 - 50m & 50m - Inf & Overall & 0 - 30m & 30 - 50m & 50m - Inf \\ \hline
HV+SV & 68.57 & 75.02 & 67.11 & 53.86 & 59.25 & 67.99 & 57.01 & 41.29 \\ \hline
DV+SV & 70.25 & 77.01 & 68.96 & 54.15 & 60.83 & 69.76 & 58.43 & 42.06 \\ \hline
MVF & \textbf{74.38} & \textbf{80.01} & \textbf{72.98} & \textbf{62.51} & \textbf{65.33} & \textbf{72.51} & \textbf{63.35} & \textbf{50.62} \\ \hline
\end{tabular}
\\
\end{center}
\caption{Comparison of methods for pedestrian detection on the Waymo Open Dataset.}
\label{eval-pedestrian}
\end{table}

\textbf{Evaluation Metrics.}
We evaluate models on the standard average precision (AP) metric for both 7-degree-of-freedom(DOF) 3D boxes and 5-DOF BEV boxes, using intersection over union (IoU) thresholds of 0.7 for vehicles and 0.5 for pedestrians, as recommended on the dataset official website.

\textbf{Experiments Setup.}
We set voxel size to $0.32$m and detection range to $[-74.88, 74.88]$m along the X and Y axes for both classes. For vehicles, we define anchors as $(l, w, h) = (4.5, 2.0, 1.6)$ m with $0$ and $\pi/2$ orientations and set the detection range to $[-5, 5]$m along the Z axis. For pedestrians, we set anchors to $(l, w, h) = (0.6, 0.8, 1.8)$ m with $0$ and $\pi/2$ orientations and set the detection range to 
 $[-3, 3]$m along the Z axis. Using the  PointPillars network backbone for both vehicles and pedestrians results in a feature map size of $468 \times 468$. 
As discussed in Section~\ref{sec:proposed_method}, pre-defining a proper setting of $K$ and $T$ for \textit{HV+SV} is critical and requires extensive experiments. Therefore, we have conducted a hyper-parameter search to choose a satisfactory configuration for this method. Here we set $K=48000$ and $T=50$ to accommodate the panoramic detection, which includes 4X more voxels and creates a 2X bigger buffer size compared to \cite{REF:pointpillars_cvpr2018}.

\textbf{Results.}
The evaluation results on vehicle and pedestrian categories are listed in Table \ref{eval-vehicle} and Table \ref{eval-pedestrian}, respectively. In addition to overall AP, we give a detailed performance breakdown for three different ranges of interest: 0-30m, 30-50m and $>$50m. We can see that \textit{DV+SV} consistently matches or improves the performance against \textit{HV+SV} on both vehicle and pedestrian detection across all ranges, which validates the effectiveness of \textit{dynamic voxelization}. Fusing multi-view information further enhances the detection performance in all cases, especially for small objects, \textit{i.e.,} pedestrians. Finally, a closer look at distance based results indicates that as the the detection range increases, the performance improvements from MVF become more pronounced. Figure~\ref{fig:error_analysis} shows two examples for both vehicle and pedestrian detection where multi-view fusion generates more accurate detections for occluded objects at long range. The experimental results also verify our hypothesis that the perspective view voxelization can capture complementary information compared to BEV, which is especially useful when the objects are far away and sparsely sampled.

\textbf{Latency.} For vehicle detection, the proposed MVF, DV+SV and HV+SV run at 65.2ms, 41.1ms and 41.1ms per frame, respectively. For pedestrian detection, the latency per frame are 60.6ms, 34.7ms and 36.1ms, for the proposed MVF, DV+SV and HV+SV, respectively.


\subsection{Evaluation on the KITTI Dataset}
\label{sec:kitti}
KITTI~\cite{REF:kitti_CVPR2012} is a popular dataset for benchmarking 3D object detectors for autonomous driving. It contains 7481 training samples and 7518 samples held-out for testing; each contains the ground truth boxes for a camera image and its associated LiDAR scan points. Similar to~\cite{REF:Multiview3D_2017}, we divide the official training LiDAR data into a training split containing 3712 samples and a validation split consisting of 3769 samples. On the derived training and validation splits, we evaluate and compare \textit{HV+SV}, \textit{DV+SV} and \textit{MVF} on the 3D vehicle detection task using the official KITTI evaluation tool. Our methods are trained with the same settings and data augmentations as in ~\cite{REF:pointpillars_cvpr2018}.

As listed in Table~\ref{eval-kitti}, using single view, \textit{dynamic voxelization} yields clearly better detection accuracy compared to \textit{hard voxelization}. With the help of multi-view information, MVF further improves the detection performance significantly. In addition, compared to other top-performing methods~\cite{REF:Multiview3D_2017,REF:VoxelNet_CVPR2018,REF:AVOD2018,REF:qi2017frustum,REF:second_2018,REF:shi2019pointrcnn}, MVF yields competitive accuracy. MFV is a general method of enriching the point level feature representations and can be applied to enhance other LiDAR-based detectors, \textit{e.g.,} PointRCNN~\cite{REF:shi2019pointrcnn}, which we plan to do in future work. 
\begin{wraptable}{r}{0.5\textwidth}
\begin{center}
\begin{tabular}{|l|c|c|c|}
\hline
\multirow{2}{*}{Method} & \multicolumn{3}{c|}{AP (IoU=0.7)}  \\ \cline{2-4}
                        & Easy & Moderate & Hard \\ \hline
MV3D~\cite{REF:Multiview3D_2017} & 71.29 & 62.68 & 56.56 \\  
VoxelNet~\cite{REF:VoxelNet_CVPR2018} & 81.98 & 65.46 & 62.85 \\ 
AVOD-FPN~\cite{REF:AVOD2018} & 84.41 & 74.44 & 68.65 \\ 
F-PointNet~\cite{REF:qi2017frustum} & 83.76 & 70.92 & 63.65 \\ 
SECOND~\cite{REF:second_2018} & 87.43 & 76.48 & 69.10 \\ 
PointRCNN~\cite{REF:shi2019pointrcnn} & 88.88 & 78.63 & \textbf{77.38} \\ 
\hline
HV+SV & 85.9 & 74.7 & 70.5 \\ 
DV+SV & 88.77 & 77.86 & 73.53 \\ 
MVF & \textbf{90.23} & \textbf{79.12} & 76.43 \\ \hline
\end{tabular} \\
\end{center}
\caption{Comparison to state of the art methods on the KITTI validation split for 3D car detection. Results of \textit{HV+SV}, \textit{DV+SV} and \textit{MVF} are based on our implementation. }
\vspace{-4em}
\label{eval-kitti}
\end{wraptable}










\vspace{-7pt}
\section{Conclusion}
\label{sec:conclusion}
We introduce MVF, a novel end-to-end multi-view fusion framework for 3D object detection from LiDAR point clouds. In contrast to existing 3D LiDAR detectors~\cite{REF:pointpillars_cvpr2018,REF:VoxelNet_CVPR2018}, which use \textit{hard voxelization}, we propose \textit{dynamic voxelization} that preserves the complete raw point cloud, yields deterministic voxel features and serves as a natural foundation for fusing information across different views. We present a multi-view fusion architecture that can encode point features with more discriminative context information extracted from the different views. Experimental results on the Waymo Open Dataset and on the KITTI dataset demonstrate that our dynamic voxelization and multi-view fusion techniques significantly improve detection accuracy. Adding camera data and temporal information are exciting future directions, which should further improve our detection framework. 

\paragraph{Acknowledgement}
We would like to thank Alireza Fathi, Yuning Chai, Brandyn White, Scott Ettinger and Charles Ruizhongtai Qi for their insightful suggestions. We also thank Yiming Chen and Paul Tsui for their Waymo Open Dataset and infrastructure-related help. 


\clearpage


\bibliography{refs}  

\begin{thebibliography}{33}
\providecommand{\natexlab}[1]{#1}
\providecommand{\url}[1]{\texttt{#1}}
\expandafter\ifx\csname urlstyle\endcsname\relax
  \providecommand{\doi}[1]{doi: #1}\else
  \providecommand{\doi}{doi: \begingroup \urlstyle{rm}\Url}\fi

\bibitem[Yang et~al.(2018)Yang, Luo, and Urtasun]{REF:yang2018pixor}
B.~Yang, W.~Luo, and R.~Urtasun.
\newblock Pixor: Real-time 3d object detection from point clouds.
\newblock In \emph{Proceedings of the IEEE Conference on Computer Vision and
  Pattern Recognition}, pages 7652--7660, 2018.

\bibitem[{Zhou} and {Tuzel}(2018)]{REF:VoxelNet_CVPR2018}
Y.~{Zhou} and O.~{Tuzel}.
\newblock Voxelnet: End-to-end learning for point cloud based 3d object
  detection.
\newblock In \emph{2018 IEEE/CVF Conference on Computer Vision and Pattern
  Recognition}, pages 4490--4499, June 2018.

\bibitem[Lang et~al.(2019)Lang, Vora, Caesar, Zhou, Yang, and
  Beijbom]{REF:pointpillars_cvpr2018}
A.~H. Lang, S.~Vora, H.~Caesar, L.~Zhou, J.~Yang, and O.~Beijbom.
\newblock Pointpillars: Fast encoders for object detection from point clouds.
\newblock \emph{CVPR}, 2019.

\bibitem[Meyer et~al.(2019)Meyer, Laddha, Kee, Vallespi-Gonzalez, and
  Wellington]{REF:lasernet_CVPR2019}
G.~P. Meyer, A.~Laddha, E.~Kee, C.~Vallespi-Gonzalez, and C.~K. Wellington.
\newblock {LaserNet}: An efficient probabilistic 3{D} object detector for
  autonomous driving.
\newblock In \emph{Proceedings of the IEEE Conference on Computer Vision and
  Pattern Recognition (CVPR)}, 2019.

\bibitem[Chen et~al.(2017)Chen, Ma, Wan, Li, and Xia]{REF:Multiview3D_2017}
X.~Chen, H.~Ma, J.~Wan, B.~Li, and T.~Xia.
\newblock Multi-view 3d object detection network for autonomous driving.
\newblock \emph{2017 IEEE Conference on Computer Vision and Pattern Recognition
  (CVPR)}, pages 6526--6534, 2017.

\bibitem[Ku et~al.(2018)Ku, Mozifian, Lee, Harakeh, and
  Waslander]{REF:AVOD2018}
J.~Ku, M.~Mozifian, J.~Lee, A.~Harakeh, and S.~Waslander.
\newblock Joint 3d proposal generation and object detection from view
  aggregation.
\newblock \emph{IROS}, 2018.

\bibitem[Sindagi et~al.(2019)Sindagi, Zhou, and
  Tuzel]{REF:MVXNetMV_Sindagi2019}
V.~A. Sindagi, Y.~Zhou, and O.~Tuzel.
\newblock Mvx-net: Multimodal voxelnet for 3d object detection.
\newblock \emph{CoRR}, abs/1904.01649, 2019.

\bibitem[Shi et~al.(2019)Shi, Wang, and Li]{REF:shi2019pointrcnn}
S.~Shi, X.~Wang, and H.~Li.
\newblock Pointrcnn: 3d object proposal generation and detection from point
  cloud.
\newblock In \emph{Proceedings of the IEEE Conference on Computer Vision and
  Pattern Recognition}, pages 770--779, 2019.

\bibitem[Girshick et~al.(2014)Girshick, Donahue, Darrell, and
  Malik]{REF:girshick2014rich}
R.~Girshick, J.~Donahue, T.~Darrell, and J.~Malik.
\newblock Rich feature hierarchies for accurate object detection and semantic
  segmentation.
\newblock In \emph{Proceedings of the IEEE conference on computer vision and
  pattern recognition}, pages 580--587, 2014.

\bibitem[Ren et~al.(2015)Ren, He, Girshick, and Sun]{REF:FasterRCNN_NIPS2015}
S.~Ren, K.~He, R.~Girshick, and J.~Sun.
\newblock Faster r-cnn: Towards real-time object detection with region proposal
  networks.
\newblock In C.~Cortes, N.~D. Lawrence, D.~D. Lee, M.~Sugiyama, and R.~Garnett,
  editors, \emph{Advances in Neural Information Processing Systems 28}, pages
  91--99. 2015.

\bibitem[{Girshick}(2015)]{REF:FastRCNN_ICCV2015}
R.~{Girshick}.
\newblock Fast r-cnn.
\newblock In \emph{2015 IEEE International Conference on Computer Vision
  (ICCV)}, pages 1440--1448, Dec 2015.

\bibitem[Redmon et~al.(2016)Redmon, Divvala, Girshick, and
  Farhadi]{REF:redmon2016you}
J.~Redmon, S.~Divvala, R.~Girshick, and A.~Farhadi.
\newblock You only look once: Unified, real-time object detection.
\newblock In \emph{Proceedings of the IEEE conference on computer vision and
  pattern recognition}, pages 779--788, 2016.

\bibitem[Liu et~al.(2016)Liu, Anguelov, Erhan, Szegedy, Reed, Fu, and
  Berg]{REF:SSD_ECCV2016}
W.~Liu, D.~Anguelov, D.~Erhan, C.~Szegedy, S.~Reed, C.-Y. Fu, and A.~C. Berg.
\newblock Ssd: Single shot multibox detector.
\newblock In B.~Leibe, J.~Matas, N.~Sebe, and M.~Welling, editors,
  \emph{Computer Vision -- ECCV 2016}, pages 21--37, Cham, 2016.

\bibitem[Redmon and Farhadi(2017)]{REF:Redmon2017YOLO9000BF}
J.~Redmon and A.~Farhadi.
\newblock Yolo9000: Better, faster, stronger.
\newblock \emph{2017 IEEE Conference on Computer Vision and Pattern Recognition
  (CVPR)}, pages 6517--6525, 2017.

\bibitem[Lin et~al.(2014)Lin, Maire, Belongie, Hays, Perona, Ramanan,
  Doll{\'a}r, and Zitnick]{REF:lin2014microsoft}
T.-Y. Lin, M.~Maire, S.~Belongie, J.~Hays, P.~Perona, D.~Ramanan,
  P.~Doll{\'a}r, and C.~L. Zitnick.
\newblock Microsoft coco: Common objects in context.
\newblock In \emph{European conference on computer vision}, pages 740--755.
  Springer, 2014.

\bibitem[{Lin} et~al.(2018){Lin}, {Goyal}, {Girshick}, {He}, and
  {Dollar}]{REF:FocalLoss_PAMI2018}
T.~{Lin}, P.~{Goyal}, R.~{Girshick}, K.~{He}, and P.~{Dollar}.
\newblock Focal loss for dense object detection.
\newblock \emph{IEEE Transactions on Pattern Analysis and Machine
  Intelligence}, pages 1--1, 2018.
\newblock ISSN 0162-8828.
\newblock \doi{10.1109/TPAMI.2018.2858826}.

\bibitem[Wang and Posner(2015)]{REF:VotingforVoting_RSS2015}
D.~Z. Wang and I.~Posner.
\newblock Voting for voting in online point cloud object detection.
\newblock In \emph{Proceedings of Robotics: Science and Systems}, Rome, Italy,
  July 2015.

\bibitem[{Engelcke} et~al.(2017){Engelcke}, {Rao}, {Wang}, {Tong}, and
  {Posner}]{REF:Vote3Deep_ICRA2017}
M.~{Engelcke}, D.~{Rao}, D.~Z. {Wang}, C.~H. {Tong}, and I.~{Posner}.
\newblock Vote3deep: Fast object detection in 3d point clouds using efficient
  convolutional neural networks.
\newblock In \emph{2017 IEEE International Conference on Robotics and
  Automation (ICRA)}, pages 1355--1361, May 2017.
\newblock \doi{10.1109/ICRA.2017.7989161}.

\bibitem[{Li}(2017)]{REF:3DFCN_RSJ2017}
B.~{Li}.
\newblock 3d fully convolutional network for vehicle detection in point cloud.
\newblock In \emph{2017 IEEE/RSJ International Conference on Intelligent Robots
  and Systems (IROS)}, pages 1513--1518, Sep. 2017.

\bibitem[Ku et~al.(2018)Ku, Mozifian, Lee, Harakeh, and
  Waslander]{REF:ku2018joint}
J.~Ku, M.~Mozifian, J.~Lee, A.~Harakeh, and S.~L. Waslander.
\newblock Joint 3d proposal generation and object detection from view
  aggregation.
\newblock In \emph{2018 IEEE/RSJ International Conference on Intelligent Robots
  and Systems (IROS)}, pages 1--8. IEEE, 2018.

\bibitem[Simony et~al.(2018)Simony, Milzy, Amendey, and
  Gross]{REF:simony2018complex}
M.~Simony, S.~Milzy, K.~Amendey, and H.-M. Gross.
\newblock Complex-yolo: an euler-region-proposal for real-time 3d object
  detection on point clouds.
\newblock In \emph{Proceedings of the European Conference on Computer Vision
  (ECCV)}, pages 0--0, 2018.

\bibitem[Qi et~al.(2017)Qi, Su, Mo, and Guibas]{REF:Qi2017PointNetDL}
C.~R. Qi, H.~Su, K.~Mo, and L.~J. Guibas.
\newblock Pointnet: Deep learning on point sets for 3d classification and
  segmentation.
\newblock \emph{2017 IEEE Conference on Computer Vision and Pattern Recognition
  (CVPR)}, pages 77--85, 2017.

\bibitem[Li et~al.()Li, Zhang, and Xia]{REF:VeloFCN2016}
B.~Li, T.~Zhang, and T.~Xia.
\newblock Vehicle detection from 3d lidar using fully convolutional network.
\newblock In \emph{RSS 2016}.

\bibitem[Qi et~al.(2017)Qi, Liu, Wu, Su, and Guibas]{REF:qi2017frustum}
C.~R. Qi, W.~Liu, C.~Wu, H.~Su, and L.~J. Guibas.
\newblock Frustum pointnets for 3d object detection from rgb-d data.
\newblock \emph{arXiv preprint arXiv:1711.08488}, 2017.

\bibitem[Xu et~al.(2018)Xu, Anguelov, and Jain]{REF:pointfusion_CVPR2018}
D.~Xu, D.~Anguelov, and A.~Jain.
\newblock {PointFusion}: Deep sensor fusion for 3d bounding box estimation.
\newblock In \emph{Proceedings of the IEEE Conference on Computer Vision and
  Pattern Recognition (CVPR)}, 2018.

\bibitem[Qi et~al.(2017)Qi, Yi, Su, and Guibas]{REF:Pointnet++NIPS2017}
C.~R. Qi, L.~Yi, H.~Su, and L.~J. Guibas.
\newblock Pointnet++: Deep hierarchical feature learning on point sets in a
  metric space.
\newblock pages 5099--5108. 2017.

\bibitem[Liang et~al.(2018)Liang, Yang, Wang, and
  Urtasun]{REF:ContFuse_ECCV2018}
M.~Liang, B.~Yang, S.~Wang, and R.~Urtasun.
\newblock Deep continuous fusion for multi-sensor 3d object detection.
\newblock In \emph{ECCV}, 2018.

\bibitem[Yang et~al.(2018)Yang, Liang, and Urtasun]{REF:HDNET_CoRL2018}
B.~Yang, M.~Liang, and R.~Urtasun.
\newblock Hdnet: Exploiting hd maps for 3d object detection.
\newblock In \emph{2nd Conference on Robot Learning (CoRL)}, 2018.

\bibitem[Liang* et~al.(2019)Liang*, Yang*, Chen, Hu, and
  Urtasun]{REF:MMF_ATG_cvpr2019}
M.~Liang*, B.~Yang*, Y.~Chen, R.~Hu, and R.~Urtasun.
\newblock Multi-task multi-sensor fusion for 3d object detection.
\newblock In \emph{CVPR}, 2019.

\bibitem[{He} et~al.(2016){He}, {Zhang}, {Ren}, and {Sun}]{REF:Resnet2016}
K.~{He}, X.~{Zhang}, S.~{Ren}, and J.~{Sun}.
\newblock Deep residual learning for image recognition.
\newblock In \emph{2016 IEEE Conference on Computer Vision and Pattern
  Recognition (CVPR)}, pages 770--778, June 2016.

\bibitem[Yan et~al.(2018)Yan, Mao, and Li]{REF:second_2018}
Y.~Yan, Y.~Mao, and B.~Li.
\newblock Second: Sparsely embedded convolutional detection.
\newblock \emph{Sensors}, 18\penalty0 (10):\penalty0 3337, 2018.

\bibitem[Kingma and Ba(2014)]{REF:Adam}
D.~P. Kingma and J.~Ba.
\newblock Adam: A method for stochastic optimization.
\newblock \emph{CoRR}, 2014.

\bibitem[{Geiger} et~al.(2012){Geiger}, {Lenz}, and
  {Urtasun}]{REF:kitti_CVPR2012}
A.~{Geiger}, P.~{Lenz}, and R.~{Urtasun}.
\newblock Are we ready for autonomous driving? the kitti vision benchmark
  suite.
\newblock In \emph{2012 IEEE Conference on Computer Vision and Pattern
  Recognition}, pages 3354--3361, June 2012.
\newblock \doi{10.1109/CVPR.2012.6248074}.

\end{thebibliography}


\end{document}